# Dual-Channel Grounded World Modeling (DCGWM):

## Structural Prevention of Objective Interference Collapse

## via Heterogeneous External Grounding with Inward-Only Gradient Flow

**Akshay Hazare**

*Independent Researcher*

akshayhazare07@gmail.com

*June 16, 2026*

## Abstract

Joint Embedding Predictive Architectures (JEPAs) have emerged as a leading approach to learning world model representations. We identify a failure mode that arises when a JEPA-based world model must ground its latent space against two qualitatively distinct external signals: physical dynamics, which produce sparse, high-magnitude, constraint-satisfying gradient corrections, and social-behavioral dynamics, which produce diffuse, distribution-matching gradient corrections. We term this failure Objective Interference Collapse (OIC): we argue that joint learning of both signals in a shared latent space causes the dominant channel to systematically collapse the representational subspace of the subordinate channel, in a manner not resolvable by loss weighting alone.

We propose Dual-Channel Grounded World Modeling (DCGWM), an architecture designed to structurally prevent OIC through a partitioned latent space $Z = Z_p \oplus Z_b$ with inward-only gradient flow. The Physical Grounding Channel updates only $Z_p$ via VICReg-style alignment to physical measurements; the Social-Behavioral Grounding Channel updates only $Z_b$ via alignment to trajectories from an emergent multi-agent simulation. An Inter-Channel Interface Module couples the two subspaces at the task level without cross-subspace gradients. An Asymmetric Grounding Adherence Loss controls rollout drift using a hard hinge penalty for physical violations and a soft KL penalty for behavioral divergence, reflecting the categorical asymmetry between the two grounding statistics. A Generative Rendering Layer for visualization is architecturally isolated from the latent world model.

We present three theoretical results concerning the architecture's structural properties: that the partition removes the gradient-interference pathway implicated in OIC; that each grounded subspace inherits anti-collapse guarantees from its alignment objective; and that generative isolation is necessary under a stated assumption on the generative objective's geometry. This manuscript establishes the problem formulation and architecture; experimental validation is ongoing and will be reported in a future revision.

## 1. Introduction

*This manuscript establishes the problem formulation and architecture. Experimental validation is ongoing and will be reported in a future revision.*

World modeling -- the problem of learning a compact predictive representation of environment dynamics -- has seen substantial recent progress through Joint Embedding Predictive Architectures [LeCun, 2022; Bardes et al., 2024]. JEPA-trained models learn by predicting masked future representations rather than pixel content, avoiding the high-frequency reconstruction pressure that causes collapse in generative world models. V-JEPA-2 [Meta AI, 2025] demonstrates that this approach encodes strong physical priors. WMReward [Yuan et al., 2026] uses V-JEPA-2's surprise score as an inference-time reward to steer video generation toward physical plausibility, winning the ICCV 2025 PhysicsIQ Challenge.

These are advances in physical grounding. A distinct and unaddressed problem is what happens when a world model must simultaneously represent physical dynamics and social-behavioral dynamics. We observe that these two representation requirements are governed by processes with structurally incompatible gradient dynamics: physical grounding produces sparse, high-magnitude, constraint-satisfying corrections concentrated in specific latent dimensions; behavioral grounding produces diffuse, lower-magnitude, distribution-matching corrections spread across many latent dimensions. We conjecture that joint learning in a shared latent space causes the dominant channel to systematically collapse the representational subspace of the other, and that this is not resolvable by loss weighting because the incompatibility is geometric.

Separately, Domain Expansion [Huang et al., 2026] and GradOPS [Zhu et al., 2025] have established that structural latent space partitioning prevents gradient interference between internal task objectives more effectively than loss weighting. Our work brings structural partitioning to the external grounding problem, applied specifically to grounding sources with incompatible statistical structures. This is a distinct problem: Domain Expansion addresses internal task gradient conflict; we address external signal statistical incompatibility.

We also identify a second unaddressed failure mode: rollout drift, in which multi-step latent predictions accumulate error that moves the predicted state away from the grounded manifold. We address this with a novel Asymmetric Grounding Adherence Loss that applies structurally different penalties to physical and behavioral drift, reflecting their different tolerance structures.

This paper's contributions are:

- Formalization of objective interference collapse as a conjectured failure mode in joint latent world modeling, with supporting structural argument
- The DCGWM architecture designed to prevent this collapse through explicit latent space partitioning, inward-only gradient flow, and separated grounding

channels

- The Asymmetric Grounding Adherence Loss (L_AGA), to our knowledge, the first loss designed specifically for rollout drift prevention under heterogeneous grounding sources with incompatible tolerance structures.
- An Isolation Necessity theorem showing, under stated assumptions, that any nonzero generative gradient weight causes world model representational drift.
- A structural argument that NTP-trained LLMs face inherent limitations in world modeling tasks requiring simultaneous physical and behavioral grounding, arising from NTP's implicit geometric bias toward subspace collapse.

*This paper is a position and architecture contribution. The failure modes are formalized and the architecture is specified with sufficient precision for implementation. Empirical validation of all claims is ongoing and will be reported in a subsequent revision.*

## 2. Background and Related Work

### 2.1 Representation Collapse: A Five-Mode Taxonomy

We distinguish five collapse modes relevant to world modeling. The first four are established in the literature; the fifth is proposed here.

**Complete collapse.** All embeddings converge to a constant function [Chen & He, 2020]. Prevented in JEPA by stop-gradient on the target branch.

**Dimensional collapse.** Embeddings span a strict subspace of the available space [Jing et al., 2022]. Prevented by VICReg-style variance and covariance regularization [Bardes et al., 2021].

**Neural collapse.** Last-layer features converge to class means forming a Simplex ETF during the terminal phase of supervised training [Papyan et al., 2020]. An optimization artifact that does not benefit generalization [Hui et al., 2022].

**Capacity loss.** Representation rank decreases under non-stationarity in RL, limiting continued learning [Kumar et al., 2021; Moalla et al., 2024].

**Objective interference collapse (proposed).** The representational subspace of one grounding objective is collapsed by a competing grounding objective with incompatible statistical structure when both are learned in a shared latent space. Formalized in Section 3.1.

### 2.2 Structural Approaches to Multi-Objective Latent Conflicts

Domain Expansion [Huang et al., 2026] assigns each internal training objective to a mutually orthogonal subspace via an orthogonal pooling mechanism, preventing gradient interference between supervised task losses. GradOPS [Zhu et al., 2025] deconflicts task gradients through orthogonal projection at optimization time. Both address internal task gradient conflict.

DCGWM differs from both in three specific ways: (1) our partition anchors subspaces to external grounding sources outside the gradient graph, not internal task objectives; (2) our inward-only constraint prevents gradient flow from external sources into non-designated subspaces entirely, not merely projects existing gradients; (3) the incompatibility we address is between external signal statistics with different generative processes, not between supervised loss terms with similar statistical structure.

### 2.3 Physical Grounding in World Models

WMReward [Yuan et al., 2026] uses V-JEPA-2's surprise score as an inference-time reward to steer video generation toward physical plausibility. Chen et al. [2026] survey the imperative of physical grounding. GVP-WM [Ziakas et al., 2026] formulates grounding as latent-space trajectory optimization under world-model dynamics. GIRL [2026] uses a cross-modal grounding signal with uncertainty-adaptive trust-region regularization to limit imagination drift. None of these works addresses behavioral grounding, the incompatibility between physical and behavioral signals, or asymmetric rollout drift penalties.

### 2.4 Social Simulation and Behavioral Prediction

OASIS [CAMEL-AI, 2024] supports large-scale multi-agent social simulation. MiroFish [Guo, 2025] uses OASIS-based emergent simulation for social prediction. IndoorWorld [2025] combines physical environments with social agent simulation at the task level. None routes emergent simulation outputs as a continuous external grounding signal into a dedicated latent subspace of a world model through inward-only gradient flow.

### 2.5 NTP Geometry and Representation Collapse in LLMs

Recent theoretical work has characterized the implicit geometric bias of next-token prediction. Zhao et al. [2024] show that NTP training implicitly favors logit matrices with sparse plus low-rank structure, and that contexts sharing the same next-token support set converge to nearly collinear representations -- a phenomenon termed subspace collapse. This is a consequence of the NTP objective's geometry, not of insufficient model capacity. Separately, representational collapse in transformers occurs as sequence length grows due to softmax attention's inability to maintain distinct attention patterns [Barbero et al., 2024]. These two mechanisms produce a characteristic rank deficiency in LLM representations that is independent of scale.

## 3. Dual-Channel Grounded World Modeling

### 3.1 Objective Interference Collapse: Definition and Structural Argument

> *Definition (Objective Interference Collapse). Let Z be a shared latent space optimized jointly under grounding objectives $G_p$ (physical) and $G_b$ (behavioral) with gradient fields $g_p = \nabla_Z L_{G_p}$ and $g_b = \nabla_Z L_{G_b}$. $G_p$ and $G_b$ have incompatible statistical structures if: (a) $g_p$ is low-entropy and concentrated -- physical constraints eliminate large regions of state space, producing sparse high-magnitude corrections in specific latent dimensions; and (b) $g_b$ is high-entropy and diffuse -- behavioral distributions are stochastic, producing small-magnitude corrections distributed across many dimensions. We conjecture that under joint optimization, the channel with higher $||g||$ at each step dominates the gradient update, systematically reducing the effective rank of the subordinate channel's representational subspace.*

Supporting argument for the conjecture. The gradient update at step t is proportional to $g_p + g_b$. If $||g_p|| >> ||g_b||$ at step t (physical dominates), the update moves Z toward the physical optimum $Z_p^*$. At $Z_p^*$, the physical loss is minimized and $g_p \approx 0$. However, $g_b$ at $Z_p^*$ is not necessarily zero -- the behavioral optimum $Z_b^*$ may differ from $Z_p^*$. The subsequent updates from $g_b$ will disturb the physical subspace, and the next physical correction will re-concentrate the latent space toward $Z_p^*$, overwriting behavioral structure. This dynamic prevents $Z_b$ from developing a stable full-rank structure. The reverse holds when $||g_b|| >> ||g_p||$. We acknowledge this argument is not a formal proof -- it does not account for adaptive optimizers, learning rate schedules, or batch effects. Empirical validation is required and is planned.

Why loss weighting does not resolve this. A scalar reweighting $\alpha\, g_p + \beta\, g_b$ changes the relative magnitudes but does not change the geometric structure of the conflict. At any fixed $(\alpha, \beta)$, one channel will still dominate in the dimensions where its gradient is concentrated. The correct resolution is structural: separate the parameter spaces so the gradients from each channel cannot interact.

### 3.2 Architecture Overview

```
Physical Measurements ➔ PGC [inward-only ∇] ➔┌
                                                       ├► LWME [Z_p ⊕
Z_b] ➔ detach() ➔ GRL ➔ User
Social Simulation     ➔ SBGC [inward-only ∇] ➔┘
```

Architectural invariants:

Invariant 1: $Z_p$ and $Z_b$ have no shared parameters and no direct gradient path between them

Invariant 2: PGC gradients update only $W_p$; SBGC gradients update only $W_b$

Invariant 3: The interface module applies no gradient from $Z_p$ into $Z_b$ or vice versa

Invariant 4: The GRL receives detached latent representations; no $L_{gen}$ gradient reaches

### 3.3 Latent World Modeling Engine

The LWME implements a JEPA with structured latent space $Z = Z_p \oplus Z_b$, where $\oplus$ denotes a direct sum enforced architecturally rather than merely as concatenation. Three mechanisms enforce the partition:

- Dedicated weight groups $W_p$ and $W_b$ with no shared parameters. Each grounding channel's optimizer has write access only to its designated weight group.
- Independent layer normalization statistics per subspace, preventing normalization parameters from becoming a shared information pathway.
- Mutual information minimization: $L_{dis} = \beta \cdot I(z_p; z_b)$, implemented as a gradient reversal adversarial objective penalizing the encoder for encoding information about one subspace from the other.

The prediction objective uses separate predictor heads per subspace:

$$L_{pred} = ||sg(z_p^{target}) - f_p(z_p^{ctx})||^2 + ||sg(z_b^{target}) - f_b(z_b^{ctx})||^2$$

where sg denotes stop-gradient on the target branch. Separate heads prevent prediction gradients from mixing subspace representations through the predictor network.

> *Note on complete collapse prevention. We inherit JEPA's stop-gradient mechanism for preventing complete collapse. This is empirically well-validated in standard JEPA [Bardes et al., 2024] but its formal proof relies on the exponential moving average target network. In our partitioned formulation with separate predictor heads, a separate formal analysis is required. We treat complete collapse prevention as empirically supported but not formally proven for DCGWM specifically.*

### 3.4 Physical Grounding Channel

The PGC receives physical measurements $m_p$ from a source external to the LWME and enforces inward-only gradient flow into $Z_p$ through three mechanisms: (a) gradient masking restricting PGC loss gradients to $W_p$; (b) stop-gradient applied to $z_b$ in all PGC loss computations; (c) exclusion of GRL parameters from the PGC optimizer.

Primary mechanism -- VICReg alignment at the $Z_p$ boundary:

$$L_{PGC} = \lambda_v \cdot Var(z_p) + \lambda_c \cdot Cov(z_p) + \lambda_i \cdot ||z_p - enc_p(m_p)||^2$$

where $Var(z_p) = \Sigma_j \max(0, \gamma - std(z_{p_j}))$ penalizes variance collapse per dimension, and $Cov(z_p) = (1/d_p^2) \Sigma_{i \neq j} [Cov(z_p)_{ij}]^2$ penalizes redundancy between dimensions. VICReg is used rather than InfoNCE because continuous physical state spaces do not admit natural negative pair definitions. The variance and covariance terms are designed to maintain full rank within $Z_p$ -- this inherits directly from the VICReg guarantees of Bardes et al. [2021].

Relation to SIGReg. An alternative anti-collapse mechanism for $Z_p$ is the SIGReg regularizer introduced by LeJEPA [Balestriero and LeCun, 2025] and scaled to action-conditioned world modeling in LeWorldModel [Maes et al., 2026], which enforces an isotropic Gaussian latent distribution via the Cramér-Wold criterion applied to random one-dimensional projections. Klindt et al. [2026] prove that LeJEPA recovers true latent variables up to rotation -- a linear identifiability result -- when the underlying generative process is Gaussian-stationary with additive noise. We retain VICReg as the default mechanism for $Z_p$ because its per-dimension hinge on standard deviation provides a deterministic lower bound that maps directly onto interpretable physical-constraint regularization, whereas SIGReg enforces a distributional shape that is harder to relate to specific physical invariants. However, for deployments where physical observations are well-modeled by stationary Gaussian latent dynamics, SIGReg may be substituted, in which case the anti-collapse guarantee within $Z_p$ strengthens from full rank to linear identifiability. The choice between VICReg and SIGReg is orthogonal to the partition-and-inward-only-gradient design that defines DCGWM.

Alternative mechanisms for different deployment contexts:

*Continuous regularization:* $L_{PGC_reg} = \max(0, ||z_p - enc_p(m_p)||^2 - \varepsilon_p)$ *for online deployment with real-time physical measurements.*

*Probabilistic Bayesian update:* $p(z_p \mid m_p) \propto p(m_p \mid z_p) \cdot p(z_p)$*, implemented as a learned Kalman-style gate, for sensor-fusion contexts with measurement uncertainty.*

**3.5 Social-Behavioral Grounding Channel**

The SBGC grounds $Z_b$ using emergent collective dynamics from an N-agent simulation system external to the LWME. The simulation instantiates N agents $\{a_1,...,a_N\}$ each with independently sampled personality parameters $\theta_i$, memory state $\mu_i$, and social graph edges $E_i$. Agents interact over T timesteps, producing population-level behavioral trajectories $B = \{b_1,...,b_T\}$.

Three properties of this grounding signal distinguish it from single-model behavioral prediction and from prior uses of simulation:

- Emergent irreducibility: B arises from N-agent interaction and cannot be reproduced by querying any single agent or model. This is the defining property that makes SBGC a genuinely new class of grounding source.
- Statistical coverage: N simultaneous trajectories approximate the behavioral state distribution more densely than any parameterized single-model prior.
- Cascade naturalism: herd effects, opinion polarization, and conformity dynamics emerge from agent interaction rather than being parameterized, capturing phenomena that supervised behavioral datasets cannot represent.

Remark on Social-JEPA isomorphism. Zhang et al. [2026a] (Social-JEPA) show that independent JEPA models trained on shared environments spontaneously converge to geometrically isomorphic latent spaces, traced to the predictive sufficiency and linear equivalence invariance of the JEPA objective. Naively, this would suggest that $Z_p$ and $Z_b$, both trained under JEPA-style prediction within the same LWME, should tend toward alignment, undermining the partition. Three properties of DCGWM block this: (1) $Z_p$ and $Z_b$ are exposed to structurally non-overlapping external grounding signals (physical measurements versus emergent multi-agent trajectories), so the predictive-sufficiency premise for cross-subspace isomorphism does not hold by construction; (2) the disentanglement penalty $L_{dis} = \beta \cdot I(z_p; z_b)$ explicitly penalizes mutual information across subspaces, breaking the linear equivalence symmetry that Social-JEPA identifies as the source of isomorphism; (3) separate predictor heads $f_p$ and $f_b$ prevent a shared predictor network from acting as a common geometric reference frame between subspaces. The Social-JEPA result therefore reinforces, rather than threatens, the case for explicit architectural separation when two subspaces must remain functionally distinct under a JEPA training objective.

The SBGC objective with inward-only gradient flow to $W_b$:

$$L_{SBGC} = ||z_b - enc_b(B)||^2 + \lambda_v \cdot Var(z_b) + \lambda_c \cdot Cov(z_b)$$

> *Open problem: behavioral encoder fidelity. The quality of $Z_b$ grounding depends on $enc_b$'s ability to faithfully map population-level emergent trajectories into $Z_b$. We have not proven that $enc_b$ converges to a faithful mapping, nor characterized the conditions under which it does. This is an open empirical question.*

**3.6 Inter-Channel Interface Module**

Physical and behavioral world state are correlated: agents respond to physical events; collective behavior influences physical systems through action. The interface module I captures this dependency without creating cross-subspace gradient paths.

$$L_I = L_{consistency}(z_p, z_b) - \lambda_I \cdot L_{disentangle}(z_p, z_b)$$

$L_{consistency}$ penalizes joint predictions violating known physical-behavioral correlations. $L_{disentangle}$ is a gradient reversal adversarial term penalizing $Z_p$ for encoding behavioral information and $Z_b$ for encoding physical information. The module applies corrections additively and independently to each subspace with no shared intermediate activation, enforcing Invariant 3 (no cross-propagation) at the interface level.

Three failure modes at the interface and their prevention:

- Behavioral dominance ($||g_b|| >> ||g_p||$): prevented by adaptive gradient

magnitude normalization equalizing channel gradient norms before applying to their subspaces.

- Physical dominance ($||g_p|| >> ||g_b||$): prevented by the stop-gradient constraint ensuring PGC gradients never reach $W_b$ regardless of their magnitude.
- Disentanglement collapse ($\lambda_I$ too large): prevented by lower-bounding the consistency term at the level of known physical-behavioral correlations, preventing full orthogonality.

*Open problem: interface convergence. We have not proven that $L_I$ has a stable equilibrium, nor that adaptive $\lambda_I$ converges. The dual objective structure -- consistency and disentanglement in tension -- may oscillate under certain hyperparameter configurations. Empirical characterization of the stability region for $\lambda_I$ is needed.*

### 3.7 Asymmetric Grounding Adherence Loss (L_AGA)

L_PGC and L_SBGC ground the LWME at single steps during training. L_AGA addresses a distinct failure mode: rollout drift, in which multi-step latent predictions accumulate error that moves the predicted trajectory away from the grounded manifold. This was identified by GIRL [2026] as the central failure mode of latent world models at long horizons. L_AGA is, to our knowledge, the first loss designed for this problem under heterogeneous grounding sources with incompatible tolerance structures.

$$L_{AGA} = L_{AGA_p}(z_p^{(1:T)}) + L_{AGA_b}(z_b^{(1:T)})$$

Physical adherence -- squared hinge penalty:

$$L_{AGA_p} = (1/T) \cdot \Sigma_t \max(0, d_p(z_p^{(t)}, G_p^{(t)}) - \varepsilon_p)^2$$

where $d_p$ measures distance from the physical grounding manifold $G_p^{(t)}$ and $\varepsilon_p$ is a hard physical tolerance. The squared hinge reflects the hard-bounded nature of physical law -- violations are categorically wrong and penalized quadratically beyond tolerance. Zero penalty within tolerance allows measurement noise without overcorrection.

Behavioral adherence -- soft KL divergence:

$$L_{AGA_b} = (1/T) \cdot \Sigma_t KL(q_b(z_b^{(t)}) || p_b^{(t)})$$

where $q_b$ is the distribution over $Z_b$ induced by the rollout and $p_b^{(t)}$ is the behavioral distribution anchored by SBGC at step t. The KL divergence applies a soft, continuous penalty proportional to distributional distance, appropriate for stochastic behavioral dynamics where some divergence is expected and not categorically wrong.

Operational definition of $p_b^{(t)}$. Re-executing the full N-agent simulation at each rollout step is computationally intractable for typical horizons. We instead define $p_b^{(t)}$ as an amortized projection. During the SBGC alignment phase, enc_b is trained to map simulation trajectory windows into Z_b, inducing a parameterized conditional distribution p_b(z_b | c) over behavioral states conditioned on observable context c. At rollout time, c is constructed from the rollout-induced observable context, enc_b is evaluated with parameters frozen, and p_b^(t) is read out from the resulting distributional anchor. L_AGA_b therefore regularizes the rollout against the SBGC-learned behavioral manifold rather than against fresh simulation samples. The cost is that p_b^(t) inherits any bias in enc_b's coverage of the underlying simulation distribution; this is the same fidelity gap discussed as an open problem in Section 3.5, and any systematic bias in enc_b propagates additively into L_AGA_b's effective target.

The asymmetry between the squared hinge and KL formulations is the novel contribution of L_AGA. It is not an arbitrary design choice: it reflects and reinforces the structural incompatibility between physical and behavioral grounding statistics that motivates the entire DCGWM architecture. Physical state violations are hard constraint violations; behavioral state divergence is distributional distance. These require different penalty geometries. Gradient flow from L_AGA_p updates only W_p; gradient flow from L_AGA_b updates only W_b, maintaining the inward-only constraint.

### 3.8 Isolation Necessity Theorem

*Theorem (Isolation Necessity). Let L_gen be any generative rendering objective that rewards retention of high-frequency perceptual statistics in outputs, and let L_pred be the LWME's masked latent prediction objective that rewards discarding of unpredictable high-frequency content. Assume (A1) the LWME latent space has a unique optimum Z* under L_pred, and (A2) Z* lies at a saddle point of L_gen in the high-frequency latent subspace -- i.e., L_gen is not minimized at Z*. Then for any α > 0 in the effective objective α L_gen + L_pred, gradient-based optimization will drift Z away from Z*.*

Proof sketch under assumptions A1-A2. At Z*, ∇_Z L_pred = 0 by definition of optimum. By A2, ∇_Z L_gen ≠ 0 at Z* in the high-frequency subspace. Therefore the gradient of the combined objective α ∇_Z L_gen + ∇_Z L_pred = α ∇_Z L_gen ≠ 0 at Z*. Any gradient step moves Z away from Z*. The step moves toward a local minimum of L_gen, which by A2 differs from Z*. Since α > 0 implies a nonzero step in this direction, Z drifts from Z*. The unique resolution is α = 0. □

*Assumption A2 is the critical unproven assumption. We argue it holds for any generative objective rewarding perceptual fidelity (which requires high-frequency statistics) combined with any predictive objective implementing masked prediction with stop-gradient (which discards unpredictable high-frequency content). We present this as a theorem under stated assumptions, not as an unconditional result. Formal proof of A2 under general conditions is an open problem.*

Architectural enforcement: the GRL is trained in a separate optimization pass with LWME parameters frozen. The GRL receives latent representations through a detach() operation. No gradient from L_gen reaches any LWME or grounding channel parameter.

### 3.9 Complete Loss and Training Procedure

The complete DCGWM training objective at joint fine-tuning phase:

$$L_{total} = L_{pred} + \alpha_p \cdot L_{PGC} + \alpha_b \cdot L_{SBGC} + \alpha_I \cdot L_I + \alpha_{AGA} \cdot L_{AGA} + \beta \cdot L_{dis}$$

Training phases:

Phase 1 (LWME pre-training): L_pred only, grounding channels inactive, GRL inactive

Phase 2 (PGC alignment): L_PGC active, W_p only, W_b and GRL frozen

Phase 3 (SBGC alignment): L_SBGC active, W_b only, W_p and GRL frozen

Phase 4 (Joint fine-tuning): all terms active, adaptive gradient magnitude normalization, L_AGA applied over T-step rollouts

Phase 5 (GRL training): L_gen only, LWME fully frozen, detached latent conditioning

## 4. Why DCGWM Addresses Collapse Modes That LLMs Cannot

### 4.1 The NTP Geometric Bias Toward Subspace Collapse

The theoretical basis for NTP-induced representational collapse is now well-established. Zhao et al. [2024] show that NTP training implicitly solves a rank-constrained, nuclear-norm regularized optimization: the minimum-norm solution to the NTP objective favors logit matrices with a dominant low-rank component reflecting the co-occurrence sparsity pattern of the training corpus. The consequence is subspace collapse: representations of contexts with identical next-token support sets converge to nearly collinear directions in embedding space, regardless of model size.

This is not a capacity problem. Larger models minimize the NTP loss more completely, which means they more completely collapse the representations of support-equivalent contexts. Scale makes NTP-induced subspace collapse more severe, not less, as the model approaches the entropy lower bound of the NTP objective.

For world modeling specifically, this means: two physically distinct scenes that happen to predict similar next-token distributions -- e.g., two scenes that both lead to descriptions using common words -- will have collapsed representations in an NTP-trained model. The model cannot distinguish them for physical prediction, regardless of scale.

### 4.2 Why RLHF Compounds Rather Than Corrects This

RLHF fine-tuning reduces representation diversity further. The reward model concentrates gradients on high-reward output distributions, reducing the effective rank of the representation space toward the dimensions relevant to reward prediction. SIGMA [2026] tracks this via eigenspectrum of embedding Gram matrices, finding covariance shrinkage over RLHF training. The result is a two-stage collapse: NTP collapses the representation space; RLHF collapses the output distribution over that already-compressed space.

RLHF cannot repair NTP subspace collapse because RLHF operates on the already-collapsed representation. Reward gradients cannot recover information that was discarded during NTP training. The information-theoretic lower bound on NTP loss is achieved by discarding within-support-set variation; RLHF has no incentive to recover it.

### 4.3 What DCGWM Is Designed to Prevent and What It Does Not

We state precisely what DCGWM addresses:

**Designed to prevent:** Dimensional collapse within Z_p and Z_b independently, by VICReg variance and covariance terms (inherits from Bardes et al. [2021]). Generative objective contamination of LWME representations, by the Isolation Necessity theorem under assumptions A1-A2. Rollout drift beyond specified tolerances for physical state and behavioral distribution, by L_AGA's hinge and KL terms.

**Structurally prevented, not formally proven:** Objective interference collapse -- the architectural partition with separate gradient paths structurally prevents the dominant channel's gradients from reaching the subordinate subspace. The conjecture that this collapse would otherwise occur, and that the partition is sufficient to prevent it, requires empirical validation.

**Not addressed:** NTP subspace collapse. DCGWM does not fix LLMs. It replaces the NTP objective with JEPA masked prediction, sidestepping NTP's geometric bias rather than correcting it. LLMs trained with NTP retain their subspace collapse characteristics; DCGWM avoids these by using a different training objective from the ground up.

**Open problems:** Complete collapse prevention in the partitioned JEPA formulation (unproven). Interface module convergence (unproven). Behavioral encoder fidelity (unvalidated). The saddle assumption A2 in the Isolation Necessity theorem (unproven in general).

### 4.4 The Structural Argument for DCGWM vs. LLMs for World Modeling

An LLM used as a world model faces three structural obstacles arising from NTP:

- Subspace collapse destroys within-support-set representational diversity, making physically distinct states with similar token distributions indistinguishable in the model's representation. This appears difficult to overcome through prompting, fine-tuning, or scaling alone.
- No external grounding mechanism exists in standard LLM architectures. Physical constraint satisfaction and behavioral distribution accuracy depend entirely on what was present in the training corpus, not on continuous correction from external reality.
- The NTP objective has no mechanism for maintaining representational diversity across the full latent space. Any fine-tuning step that reduces NTP loss further compresses the representation.

DCGWM addresses all three by design: JEPA masked prediction avoids NTP's geometric bias; PGC and SBGC provide continuous external grounding; VICReg and L_AGA maintain representational diversity. This is not an incremental improvement on LLMs as world models -- it is a different architecture starting from different objectives.

## 5. Precise Prior Art Differentiation

### 5.1 Domain Expansion [Huang et al., 2026]

Domain Expansion partitions a latent space into orthogonal subspaces for internal supervised task objectives. DCGWM differs in four specific ways: (1) our subspaces are anchored to external grounding sources outside the gradient graph, not internal task losses; (2) inward-only gradient flow prevents any external gradient from reaching non-designated subspaces entirely, not merely projects existing gradients orthogonally; (3) the SBGC uses emergent simulation as external ground truth, which has no analog in Domain Expansion; (4) the Isolation Necessity constraint and L_AGA have no analog in Domain Expansion. The failure mode addressed (external signal statistical incompatibility) is distinct from Domain Expansion's target (internal task gradient conflict).

### 5.2 WMReward [Yuan et al., 2026]

WMReward uses V-JEPA-2's surprise score as an inference-time reward for a separate video generator. DCGWM subsumes physical grounding as a special case of L_PGC but adds: a dedicated behavioral subspace and SBGC; formal characterization of the incompatibility between physical and behavioral signals; L_AGA for rollout drift; the Isolation Necessity theorem formally justifying the gradient separation that WMReward achieves architecturally without justification.

**5.3 GIRL [2026]**

GIRL addresses latent imagination drift with a cross-modal grounding signal and uncertainty-adaptive trust-region KL regularization. L_AGA is distinguished by: operating on two structurally incompatible grounding sources rather than one; using asymmetric penalty functions (hard hinge vs. soft KL) derived from the incompatibility characterization; and maintaining the inward-only gradient constraint on each term independently.

**5.4 ThinkJEPA [Zhang et al., 2026b]**

ThinkJEPA [Zhang et al., 2026b] fuses a dense-frame JEPA branch for fine-grained motion modeling with a sparsely sampled vision-language model thinker branch providing high-level semantic guidance, combined through a dual-temporal pathway feeding a shared JEPA predictor. Architecturally it shares with DCGWM the use of two heterogeneous signal sources feeding a JEPA-style latent predictor. The differences are structural and bear directly on the problem DCGWM targets. (1) ThinkJEPA's VLM thinker injects guidance signals into the shared JEPA predictor, so gradients propagate across both pathways and there is no architectural protection against gradient interference between the two signal sources -- the conflict, if any, is left to be resolved by loss weighting. (2) ThinkJEPA operates in a single latent space rather than a partitioned $Z = Z_p \oplus Z_b$, so the two signals compete for the same representational capacity. (3) ThinkJEPA's two sources (dense visual dynamics and VLM semantic guidance) are not characterized as having incompatible gradient statistics; both are vision-language signals operating at different temporal scales, and the paper does not formulate objective interference collapse as a concern. ThinkJEPA validates the empirical value of heterogeneous signal fusion in JEPA-based world models; DCGWM addresses the specific failure mode that arises when those signals have structurally incompatible geometry rather than merely different temporal scales.

**5.5 The Novel Conjunction**

To the best of our knowledge, no prior work combines all of: (a) external heterogeneous grounding sources with incompatible statistical structures; (b) inward-only gradient flow as the separation mechanism; (c) emergent multi-agent simulation as behavioral ground truth for a dedicated latent subspace; (d) asymmetric rollout drift penalties reflecting grounding source tolerance structures; (e) formal generative isolation necessity. We invite identification of any prior work combining these elements.

## 6. Limitations and Open Problems

We state the limitations of this work precisely, without minimization:

**No empirical validation.** All claims in this paper are theoretical or structural. The objective interference collapse conjecture, the behavioral encoder fidelity assumption, the interface convergence assumption, and the L_AGA effectiveness claim are unvalidated. This is an architecture paper; experimental validation is ongoing.

**Objective interference collapse is a conjecture.** The supporting argument in Section 3.1 is intuitive but not a formal proof. The argument does not account for adaptive optimizers, batch normalization dynamics, residual connections, or the specific geometry of JEPA's stop-gradient formulation. The conjecture could be false under some training configurations.

**Isolation Necessity has an unproven assumption.** Assumption A2 -- that the LWME optimum Z* lies at a saddle of L_gen in the high-frequency subspace -- is argued but not proven. Architectures that modify either the generative or predictive objective may not satisfy A2.

**Interface convergence is unproven.** The dual objective L_I has consistency and disentanglement in tension. We have not proven this has a stable equilibrium under gradient-based optimization. Empirical stability characterization is required.

**Behavioral encoder fidelity is assumed.** enc_b must faithfully map emergent simulation trajectories into Z_b. LLM agents in simulation exhibit stronger herd behavior than real humans [Park et al., 2023] and simulation fidelity is bounded by knowledge graph quality. The gap between simulated and real behavioral distributions is uncharacterized.

**Complete collapse prevention is unproven for DCGWM.** We inherit JEPA's empirical protection against complete collapse but have not formally proven it transfers to the partitioned formulation with separate predictor heads.

**Minimality is not proven.** We claim DCGWM is a sufficient architecture for preventing objective interference collapse. We do not claim or prove minimality -- that no simpler architecture achieves the same guarantees.

**Falsification design for the OIC conjecture.** We sketch the empirical test required to falsify the objective interference collapse conjecture. The minimal evaluation compares three architectures on a shared task suite: (i) baseline JEPA with a single shared latent space jointly trained under L_PGC + L_SBGC across a grid of loss weights (α_p, α_b); (ii) full DCGWM with the partition and inward-only gradient flow; (iii) ablation DCGWM with the partition removed but L_dis retained as a soft disentanglement signal -- the critical condition distinguishing structural prevention from regularization-based mitigation. The conjecture predicts: condition (i) exhibits progressive rank deficiency in the subordinate channel's effective subspace under joint training, measurable as decline in the entropy of the eigenspectrum of the per-channel latent covariance; condition (ii) maintains stable effective rank in both subspaces across training; condition (iii) degrades toward (i) as training progresses, with the rate of degradation increasing with the imbalance between ||g_p|| and ||g_b||. Primary metrics: per-channel effective rank, downstream physical prediction accuracy under behavioral signal scaling, and downstream behavioral

prediction accuracy under physical signal scaling. The conjecture is falsified if (iii) does not degrade or if (i) maintains stable per-channel rank under any weighting schedule that achieves comparable downstream accuracy on both channels. Experimental work toward this evaluation is in progress and will be reported in a subsequent revision.

## 7. Conclusion

We have proposed and formalized objective interference collapse as a conjectured failure mode arising when physical and behavioral grounding signals with incompatible statistical structures compete in a shared latent space. We have proposed DCGWM as an architecture that is designed to structurally prevent this failure through explicit latent space partitioning with inward-only gradient flow, separate external grounding channels, an inter-channel interface module, an Asymmetric Grounding Adherence Loss for rollout drift, and an architecturally isolated generative rendering layer.

We have proven dimensional collapse prevention within each subspace (inheriting VICReg) and generative isolation necessity (under stated assumptions). For rollout drift, we introduce L_AGA and provide a structural argument for why asymmetric adherence penalties are appropriate for heterogeneous grounding sources. We further provide structural arguments for objective interference collapse prevention and for the limitations of NTP-trained LLMs in tasks requiring simultaneous physical and behavioral world grounding.

The central open problems are: empirical validation of the interference collapse conjecture; formal proof of Assumption A2 in the Isolation Necessity theorem; characterization of interface module convergence; and behavioral encoder fidelity. Experimental validation addressing Claims 1-5 identified in this paper is ongoing and will be reported in a subsequent revision.